\documentclass[compsoc, conference, letterpaper, 10pt, times]{IEEEtran}

\IEEEspecialpapernotice{(Invited Paper)}
\IEEEoverridecommandlockouts

\usepackage{amssymb}
\usepackage{amsfonts}
\usepackage{url}
\usepackage{graphicx}

\usepackage{flushend}

\title{On the Protection of Private Information in Machine Learning Systems:\\ Two Recent Approaches}
\author{\IEEEauthorblockN{Mart{\'\i}n Abadi,
    \'Ulfar Erlingsson,
    Ian Goodfellow,
    H.~Brendan McMahan,\\
    Ilya Mironov,
    Nicolas Papernot$^\ast$,\thanks{\rm $^\ast$~Nicolas Papernot is an intern at Google; he is a student at Pennsylvania State University.}
    Kunal Talwar, and Li Zhang}
\IEEEauthorblockA{Google}
  }

\begin{document}
\maketitle
\begin{abstract}
  The recent, remarkable growth of machine learning has led to intense
  interest in the privacy of the data on which machine learning
  relies, and to new techniques for preserving privacy. However, older
  ideas about privacy may well remain valid and useful.  This note
  reviews two recent works on privacy in the light of the wisdom of
  some of the early literature, in particular the principles distilled
  by Saltzer and Schroeder in the 1970s.
\end{abstract}

\section{Introduction}

In their classic tutorial paper, Saltzer and Schroeder described the
mechanics of protecting information in computer systems, as it was
understood in the mid 1970s~\cite{DBLP:journals/pieee/SaltzerS75}.  They were
interested, in particular, in mechanisms for achieving privacy, which
they defined as follows:
\begin{quote}
The term ``privacy'' denotes a socially defined ability of an
individual (or organization) to determine whether, when, and to whom
personal (or organizational) information is to be released.
\end{quote}
Saltzer and Schroeder took ``security'' to refer to the body of
techniques for controlling the use or modification of computers or
information.  In this sense, security is an essential element of
guaranteeing privacy. Their definitions are roughly in line with our
current ideas, perhaps because they helped shaped those ideas. (In
contrast, other early definitions took ``security'' to refer to the
handling of classified information, more narrowly~\cite{Ware:1967:SPS:1465482.1465525}.)

Although some of that early literature may appear obsolete (for
example, when it give statistics on computer abuse~\cite[Table~1]{turnware75}),
it makes insightful points that remain valid today. In particular,
Ware noted that ``one cannot exploit the good will of users as
part of a privacy system's design''~\cite{Ware:1967:SPS:1465482.1465525}.
Further, in their discussion of
techniques for ensuring privacy~\cite[p.~9]{turnware75}, Turn and Ware
mentioned randomized response---several decades before the RAPPOR
system~\cite{DBLP:conf/ccs/ErlingssonPK14} made randomized response commonplace on the Web.

The goal of this note is to review two recent works on
privacy~\cite{DBLP:conf/ccs/AbadiCGMMT016,DBLP:journals/corr/PapernotAEGT16}
in the light of the wisdom of some of the early literature. Those two
works concern the difficult problem of guaranteeing privacy properties
for the training data of machine learning systems. They are
particularly motivated by contemporary deep
learning~\cite{deeplearning,Goodfellow-et-al-2016-Book}. They employ two very different
techniques: noisy stochastic gradient descent (noisy SGD) and private
aggregation of teacher ensembles (PATE).  However, they both rely on a
rigorous definition of privacy, namely differential
privacy~\cite{DBLP:conf/tcc/DworkMNS06,Dwork-CACM}.  The growing body
of research on machine learning with privacy includes several
techniques and results related to those we review
(e.g.,~\cite{KasiviswanathanLNRS11,chaudhuri2011,KST12,SongCS13,BassilyTS14,ShokriShmatikov15,DBLP:conf/icml/HammCB16,WCJN16-RDBMS}), and
undoubtedly neither noisy SGD nor PATE constitutes the last word on the subject;
surveying this research is beyond the scope of this note.

Saltzer and Schroeder observed that no complete method existed for
avoiding flaws in general-purpose systems. Despite much progress, both
in attacks and in defenses, their observation is still correct; in
particular, neither the concept of differential privacy nor the
sophisticated techniques for achieving differential privacy are a
panacea in this respect.  Indeed, they may even present opportunities
for new flaws~\cite{Mironov:2012:SLS:2382196.2382264}.

As a mitigation, Saltzer and Schroeder identified several useful
principles for the construction of secure systems.  In this note, we
discuss how those principles apply (or fail to apply) to modern
systems that aim to protect private information. Specifically, we
review those two recent works on protecting the privacy of training
data for machine learning systems, and comment on their suitability by
the standards of those principles. Accordingly, our title echoes
Saltzer and Schroeder's, ``The Protection of Information in Computer
Systems''. 

In the next section, we describe the problem of interest in more
detail. In Sections~\ref{sec:sgd} and~\ref{sec:pate} we review the
works on noisy SGD and PATE, respectively. In Section~\ref{sec:prin}
we discuss these works, referring to Saltzer and Schroeder's
principles.  We conclude in Section~\ref{sec:conc}.

\section{The Problem}\label{sec:problem}

Next, in Section~\ref{sec:framing}, we describe the problem on which
we focus, at a high level. In Section~\ref{sec:otherpbs}, we describe
related problems that are not addressed by the techniques that we
discuss in the remainder of the note.

\subsection{Framing}\label{sec:framing}

We are broadly interested in supervised learning of classification
tasks. A classification task is simply a function $f$ from examples to
classes, for instance from images of digits to the corresponding
integers. (In a more refined definition, $f$
may assign a probability to each class for each example; we omit these probabilities below.)
The learning of such a task means finding another function
$g$, called a model, that approximates~$f$ well by some metric. Once
the model $g$ is picked, applying it to inputs is called inference.  The
learning is supervised when it is based on a collection of known
input-output pairs (possibly with some errors); this collection is the training data.

Since this training data may be sensitive, its protection is an
obvious concern, but the corresponding threat model is somewhat less
obvious. Attacks may have at least two distinct goals, illustrated by the work of Fredrikson et al.~\cite{FJR15} and that of Shokri et al.~\cite{DBLP:journals/corr/ShokriSS16}, respectively:
\begin{itemize}
\item the extraction of training data (total or partial) from a model $g$, or
\item testing whether an input-output pair, or simply an input or an output, is part of the training data.
\end{itemize}
We wish to prevent both.  The definition of differential privacy gives
bounds on the probability that two datasets can be distinguished, thus rigorously
addressing membership tests. The extraction of training data seems a
little difficult to characterize formally; intuitively, however, it
appears harder than membership tests.
(Shokri et al.~also discuss a weak form of model inversion that appears incomparable with membership tests.)
Therefore, we focus on membership tests, and rely on the definition of differential privacy.

Moreover, at least two kinds of threats are worth considering; we call them ``black-box'' and ``white-box'', respectively:
\begin{itemize}
\item ``black-box'': attackers can apply the model $g$ to new inputs of their choice, possibly up to some number of times or under other restrictions, or
\item ``white-box'': attackers can inspect the internals of the model $g$.
\end{itemize}
``White-box'' threats subsume ``black-box'' threats (in other words,
they are more severe), since attackers with access to the internals of
a model can trivially apply the model to inputs of their choice.
``Black-box'' threats may be the most common, but full-blown ``white-box''
threats are not always unrealistic, in particular if models are deployed on
devices that attackers control.  Finally, the definition of ``white-box''
threats is simpler than that of ``black-box'' threats, since it does
not refer to restrictions.  For these reasons, we focus on
``white-box'' threats.

Attackers may also act during the learning process, for example
tampering with some of the training data, or reading intermediate
states of the learning system.  Noisy SGD and PATE are rather
resilient to those attacker capabilities, which we do not consider in
detail for simplicity.

The term ``protection'' sometimes refers, specifically, to protection
from programs~\cite{DBLP:journals/sigops/Lampson74}. When we discuss
the protection of training data, we may use the term with a broader,
informal meaning. This distinction is unimportant for our
purposes. 

\subsection{Other Problems}\label{sec:otherpbs}

Related problems pertain to other learning tasks, to other data, and
to other aspects of machine learning systems.

\subsubsection{Privacy in Other Learning Tasks}

There is more to machine learning than the supervised learning of
classification tasks. In particular, generative models, which aim to
generate samples from a distribution, are another important province
of machine learning.

Thinking about privacy has not progressed at an
even pace across all areas of machine learning. However, we may hope
that some core ideas and techniques may be broadly applicable. For example, many learning techniques employ SGD-like iterative techniques, and noisy variants of those may be able to guarantee privacy properties.

\subsubsection{Privacy for Inference Inputs}

This note focuses on training data, rather than on 
the inputs that machine learning systems receive after they are
trained and deployed, at inference time. Cryptographic techniques, such as 
such as
CryptoNets~\cite{DBLP:conf/icml/Gilad-BachrachD16},
can protect those inputs.

Technically, from a privacy perspective, training time and inference
time are different in several ways. In particular, the ability
of machine learning systems to memorize
data~\cite{DBLP:journals/corr/ZhangBHRV16} is a concern at training
time but not at inference time. Similarly, it may be desirable for a machine learning system to
accumulate training data and to use the resulting models for some
time, within the bounds of data-retention policies, while the system may
not need to store the inputs that it receives for inference.

Finally, users's concerns about privacy may be rather different with
regards to training and inference.  While in many applications users
see a direct, immediate benefit at inference time, the connection
between their providing data and a benefit to them or to society (such
as improved service) is less evident at training time. Accordingly, some users may
well be comfortable submitting their data for inference but not contributing it for training purposes.

Nevertheless, some
common concerns about privacy focus on inference rather than training.
The question ``what can Big Brother infer about me?'' pertains to
inference. It is outside the scope of the techniques that we discuss
below.

\subsubsection{A Systems Perspectives}\label{sec:systems}

Beyond the core of machine learning algorithms, privacy may depend on 
other aspects of the handling of
training data in machine learning systems, throughout the
data's life cycle:
\begin{itemize}
\item measures for
sanitizing the data, such as anon\-y\-mization, pseudonymization,
aggregation, generalization, and the stripping of outliers, when the
data is collected;
\item traditional access controls, for the raw data after its collection
  and perhaps also for derived data and the resulting machine learning models; and
\item finally, policies for data retention and mechanisms for data deletion.
\end{itemize}
In practice, a holistic view of the handling of private information is essential for providing meaningful
end-to-end protection.

\section{Noisy SGD}\label{sec:sgd}

Many machine learning techniques rely on parametric functions as
models. Such a parametric function $g$ takes as input a parameter $\theta$ and an
example $x$ and outputs a class $g(\theta,x)$.
For instance, $\theta$ may be the collection of weights and biases of a deep neural network~\cite{deeplearning,Goodfellow-et-al-2016-Book}.
With each $g$ and $\theta$ one associates
a loss $L(g,\theta)$, a value that quantifies the cost of any
discrepancies between the model's prediction $g(\theta,x)$ and the
true value $f(x)$, over all examples~$x$. The loss over the true
distribution of examples~$x$ is approximated by the loss over the
examples in the training data and, for those, one takes $f(x)$ to be as
given by the training data, even though this training data may
occasionally be incorrect. Training the model $g$ is the process of
searching for a value of $\theta$ with the smallest loss
$L(g,\theta)$, or with a tolerably small loss---global minima are
seldom guaranteed. After training, $\theta$ is fixed, and new examples
can be submitted. Inference consists in applying $g$ for a fixed value
of $\theta$.

Often, both the model $g$ and the loss $L$ are differentiable
functions of~$\theta$. Therefore, training often relies on gradient
descent. With SGD, one repeatedly picks an example $x$ (or a
mini-batch of such examples), calculates $g(\theta,x)$ and the
corresponding loss for the current value of $\theta$, and adjusts
$\theta$ in order to reduce the loss by going in the opposite
direction of the gradient. The magnitude of the adjustment depends on
the chosen learning rate.

The addition of noise is a common technique for achieving privacy
(e.g.,~\cite{turnware75}), and also a common technique in deep
learning (e.g.,~\cite{NVLSK15}), but for privacy purposes the noise
should be carefully calibrated~\cite{DBLP:conf/tcc/DworkMNS06}.  The
sensitivity of the final value of $\theta$ to the elements of the
training data is generally hard to analyze. On the other hand, since
the training data affects $\theta$ only via the gradient computations,
we may achieve privacy by bounding gradients (by clipping) and by
adding noise to those computations.  This idea has been developed in
several algorithms and systems
(e.g.,~\cite{SongCS13,BassilyTS14,ShokriShmatikov15}).  Noisy SGD, as
defined in~\cite{DBLP:conf/ccs/AbadiCGMMT016}, is a recent embodiment
of this idea, with several modifications and extensions, in particular
in the accounting of the privacy loss.

\section{PATE}\label{sec:pate}

The use of ensembles of models is common in machine
learning~\cite{dietterich2000ensemble}. If an ensemble comprises a
large enough number of models, and each of the models is trained with
a disjoint subset of the training data, we may reason, informally, that any
predictions made by most of the models should not be based on
any particular piece of the training data. In this sense, the
aggregation of the models should protect privacy with respect to
``black-box'' threats.

Still, since the internals of each of the models in an ensemble is
derived from the training data, their exposure could compromise
privacy with respect to ``white-box'' threats.
In order to overcome this
difficulty, we may treat the ensemble as a set of ``teachers'' for a
new ``student'' model. The ``student'' relies on the ``teachers''
only via their prediction capabilities, without access to their
internals.

Training the ``student'' relies on querying the ``teachers'' about
unlabelled examples. These should be disjoint from the training data
whose privacy we wish to protect.  It is therefore required that such
unlabelled examples be available, or that they can be easily
constructed.  The queries should make efficient use of the
``teachers'', in order to minimize the privacy cost of these queries.
Once the ``student'' is fully trained, however, the ``teachers'' (and any secrets they keep) can be discarded.

PATE and its variant PATE-G are based on this strategy. They belong in
a line of work on knowledge aggregation and transfer for
privacy~\cite{nissim2007smooth,pathak2010multiparty,DBLP:conf/icml/HammCB16}.
Within this line of work, there is some diversity in goals and in
specific techniques for knowledge aggregation and transfer.  PATE aims
to be flexible on the kinds of models it supports, and is applicable,
in particular, to deep neural networks. It relies on noisy plurality
for aggregation; the noise makes
it possible to derive
differential-privacy guarantees. The variant PATE-G relies on
generative, semi-supervised methods for the knowledge transfer;
currently, techniques based on generative adversarial
networks (GANs)~\cite{goodfellow2014generative,salimans2016improved} are
yielding the best results.  They lead, in particular, to
state-of-the-art privacy/utility trade-offs on MNIST and SVHN
benchmarks.

\section{Principles, Revisited}\label{sec:prin}

In this section, we consider how the principles distilled by Saltzer
and Schroeder apply to noisy SGD and to PATE.  Those principles have
occasionally been analyzed, revised, and extended over the years
(e.g.,~\cite{Saltzer:2009:PCS:1594884,Garfinkel:2005:DPP:1195191,Smith:2012:CLS:2420631.2420856}). We
mainly refer to the principles in their original form.

Most of the protection mechanisms that Saltzer and Schroeder discussed
are rather different from ours in that they do not involve data
transformations. However, they also described protection
by encryption (a ``current research direction'' when they wrote their
paper); the training of a model from data is loosely analogous to
applying a cryptographic transformation to the data. This analogy
supports the view that noisy SGD and PATE should be within the scope of
their principles.

\subsubsection*{Economy of mechanism}
This principle says that the design of protection mechanisms should be
kept as simple and small as possible.

Neither noisy SGD nor PATE (and its variants)
seem to excel in this respect, though for somewhat different reasons:
\begin{itemize}
\item Although noisy SGD relies on simple algorithmic ideas that can be
implemented concisely, these ideas directly affect SGD, which is so
central to many learning algorithms. Therefore, applying noisy SGD is
akin to performing open-heart surgery. The protection mechanism is not
a stand-alone system component. Moreover, optimizations and extensions
of learning algorithms (for example, the
introduction of techniques such as batch
normalization~\cite{DBLP:conf/icml/IoffeS15}) may require new methods
or new analysis.

\item PATE involves somewhat more design details than noisy SGD. In
particular, PATE-G incorporates sophisticated techniques based on GANs.
On the other hand, those design details are
entirely separate from the training of the ``teacher'' models, and
independent of the internal structure of the ``student'' model.
\end{itemize}
It remains to be seen whether radically simpler and smaller mechanisms
can be found for the same purpose.

\subsubsection*{Fail-safe defaults} This principle means that lack of access is the default. In particular, mistakes should result in refusing permission.

This principle, which is easy to interpret for traditional reference
monitors, appears difficult to apply to noisy SGD, PATE, and other
techniques with similar goals, which generally grant the same level of
access to anyone making a request under any circumstances.

We note, however, that many of these techniques achieve a guarantee
known as $(\epsilon,\delta)$-differential-privacy~\cite{ODO};
while the parameter $\epsilon$
from the original definition of differential privacy can be viewed
as a privacy cost~\cite{DBLP:conf/tcc/DworkMNS06},
the additional parameter $\delta$ can be viewed as a probability of failure.
Prima facie, such a failure results in a loss of privacy, rather than
a loss of accuracy. In this sense, the guarantee does not imply
fail-safe defaults. A more refined analysis paints a subtle and
interesting picture with a trade-off between the parameters $\epsilon$ and $\delta$~\cite{DBLP:journals/corr/Mironov17}.

\subsubsection*{Complete mediation} This principle implies that every access to sensitive data should go through the protection mechanism.

Resistance to ``white-box'' threats means that the
internals of models are not sensitive, so concerns about complete
mediation should not apply to them, but these concerns still apply to the raw
training data at rest or in transit.
Complete mediation requires a system-wide perspective
(see Section~\ref{sec:systems}).

\subsubsection*{Open design} This principle, which echoes one of
Kerckhoffs's~\cite{kerckhoffs1883cryptographieone}, states that the designs of
protection mechanisms should not depend on secrecy, and should not be kept secret.

Both noisy SGD and PATE are satisfactory in this respect. This
property may seem trivial until one notes that not all current work on
privacy (and, in particular, on differential privacy) is equally open.

\subsubsection*{Separation of privilege} This principle calls for the use of multiple independent ``keys'' for unlocking access.

Like the principle of fail-safe defaults, it appears difficult to
apply to noisy SGD and to PATE. It may perhaps apply to a separate,
outer level of protection.

\subsubsection*{Least privilege} This principle reads ``Every program and every user of the system should operate using the least set of privileges necessary to complete the job''.

This principle seems more pertinent to the implementations of noisy
SGD and PATE than to their high-level designs.  For instance, in the
case of PATE, it implies that each of the ``teacher'' models should be
configured in such a way that it would not have access to the training
data of the others, even if its software has flaws.

The principle has the virtue of limiting the damage that may be caused
by an accident or error. We simply do not have enough experience with
noisy SGD and PATE to characterize the nature and frequency of accidents
and errors, but it seems prudent to admit that they are possible, and to
act accordingly.

\subsubsection*{Least common mechanism} This principle addresses the difficulty of providing mechanisms shared by more than one user. Those mechanisms may introduce unintended communication channels. Moreover, it may be hard to satisfy all users with any one mechanism.

The principle can be regarded as an end-to-end
argument~\cite{Saltzer:2009:PCS:1594884}, since it suggests that
shared mechanisms should not attempt that which can be achieved by
each user separately. While each user could ensure the differential
privacy of their data by adding noise, as in
RAPPOR~\cite{DBLP:conf/ccs/ErlingssonPK14}, the required levels of
noise can sometimes conflict with utility.  Therefore, shared
mechanisms that achieve differential privacy are attractive.

In techniques such as noisy SGD and PATE, the privacy parameters (in
particular the parameters $\epsilon$ and $\delta$ discussed above) are
the same for all pieces of the training data, and for all accesses to
the learning machinery.  The addition of
weights~\cite{Proserpio:2014:CDS:2732296.2732300} could perhaps
accommodate the privacy requirements of different pieces of training
data, and thus those of different users.

\subsubsection*{Psychological acceptability} This principle advocates ease of use, and was later called the principle of least astonishment~\cite{Saltzer:2009:PCS:1594884}. Saltzer and Schroeder noted: ``to the extent that the user's mental image
of his protection goals matches the mechanisms he must use, mistakes
will be minimized''.

This remark by Saltzer and Schroeder was in fact one of the starting
points for the work on PATE. While noisy SGD provides mathematical
guarantees, understanding them requires a fair amount of
sophistication in differential privacy and in machine learning, which
many users and operators of systems will lack. In contrast, the way in
which PATE guarantees privacy should be intuitively clear, at least at
a high level. No advanced background is required in order to accept
that, if 100 independently trained machine learning models say that a
picture is an image of a cat, then this prediction is probably
independent of any particular picture in their disjoint sets of
training data.

\subsubsection*{Work factor} This principle calls for comparing the resources of attackers with 
the cost of circumventing the protection mechanism.

The ``white-box'' threat model is helpful in simplifying work-factor
considerations. The privacy guarantees
for noisy SGD and PATE benefit from the fact that attackers need not be limited to any
particular number of queries at inference time.

\subsubsection*{Compromise recording} The final principle suggests that it is advantageous to be able to detect and report any failures of protection.

As noted above, $(\epsilon,\delta)$-differential-privacy includes the
possibility of failures. In our setting, it allows for the possibility
that noisy SGD and PATE plainly reveal one or a few pieces of the training
data that they should safeguard. No detection or reporting of such failures has been
contemplated. The seriousness of this problem seems open to debate, as
it may be a shortcoming of the theory, rather than of the algorithms.

\section{Conclusion}\label{sec:conc}

The current, vibrant research on privacy is developing sophisticated
concepts and techniques that apply, in particular, to cutting-edge
machine learning systems. Noisy SGD and PATE, which this note reviews,
aim to contribute to this line of work. Looking beyond the core
algorithms, it is important to understand how these algorithms fit
into systems and society. Economy of mechanism, psychological
acceptability, and other fundamental principles should continue to
inform the design and analysis of the machinery that aims to protect
privacy.

\subsection*{Acknowledgments}
We are grateful to Mike Schroeder
for discussions on the matter of this note and for comments on a draft.

\bibliographystyle{IEEEtran}
\bibliography{csf17}

\end{document}